\RequirePackage{fix-cm}
\documentclass[smallextended]{svjour3}       
\smartqed  
\usepackage{booktabs}
\usepackage{graphicx}
\usepackage[round]{natbib}
\usepackage{algorithm}
\usepackage{algpseudocode}
\usepackage{tablefootnote}
\usepackage{multirow}
\usepackage{framed}
\usepackage{hyperref}
\usepackage{amsfonts} 
\usepackage{bbm}
\usepackage{newfloat}
\usepackage{listings}
\usepackage[normalem]{ulem}
\usepackage[table,dvipsnames]{xcolor}
\usepackage{hhline}
\usepackage{tcolorbox}
\usepackage{enumitem}
\usepackage{makecell}

\newcolumntype{P}[1]{>{\centering\arraybackslash}m{#1}}

\usepackage{soul}

\begin{document}

\title{DiverGet: A Search-Based Software Testing Approach for Deep Neural Network Quantization Assessment
}


\author{Ahmed Haj Yahmed        \and Houssem Ben Braiek \and Foutse Khomh \and Sonia Bouzidi \and Rania Zaatour
}


\institute{Ahmed Haj Yahmed \at
              \email{ahmed.haj-yahmed@polymtl.ca}
}

\date{Received: date / Accepted: date}

\maketitle

\begin{abstract}
Quantization is one of the most applied Deep Neural Network (DNN) compression strategies, when deploying a trained DNN model on an embedded system or a cell phone. This is owing to its simplicity and adaptability to a wide range of applications and circumstances, as opposed to specific Artificial Intelligence (AI) accelerators and compilers that are often designed only for certain specific hardware (e.g., Google Coral Edge TPU). With the growing demand for quantization, ensuring the reliability of this strategy is becoming a critical challenge. Traditional testing methods, which gather more and more genuine data for better assessment, are often not practical because of the large size of the input space and the high similarity between the original DNN and its quantized counterpart. As a result, advanced assessment strategies have become of paramount importance. In this paper, we present DiverGet, a search-based testing framework for quantization assessment. DiverGet defines a space of metamorphic relations that simulate naturally-occurring distortions on the inputs. Then, it optimally explores these relations to reveal the disagreements among DNNs of different arithmetic precision. We evaluate the performance of DiverGet on state-of-the-art DNNs applied to hyperspectral remote sensing images. We chose the remote sensing DNNs as they're being increasingly deployed at the edge (e.g., high-lift drones) in critical domains like climate change research and astronomy. Our results show that DiverGet successfully challenges the robustness of established quantization techniques against naturally-occurring shifted data, and outperforms its most recent concurrent, DiffChaser, with a success rate that is (on average) four times higher. 

\end{abstract}

\section{Introduction}
\label{intro}
After the introduction of the ImageNet dataset~\citep{deng_imagenet:_2009}, Deep Neural Networks (DNNs) have gotten a lot of attention from academia and industry. In fact, over the last years, DNNs achieved state-of-the-art results in many tasks, such as image recognition~\citep{he_deep_2015}, natural language processing~\citep{young_recent_2018}, and speech recognition~\citep{hinton_deep_2012}. 

Recently, there has been a constant movement of Artificial Intelligence (AI) towards edge devices. The on-device inference is the process of making predictions using a trained model that is running on the device. This trend offers numerous advantages, mainly the decrease in inference latency. In fact, by bypassing the data upload to the server and the wait time for the inference result, the application is able to respond to the user's request faster. Besides, discarding the server reliance has other benefits, like being able to operate with little to no connectivity, and lowering privacy concerns as the user's data remain on the device. However, one limitation that hinders DNN models from being widely deployed on edge is their resources consumption. In fact, they require a lot of memory and might compromise the battery life of devices during training and inference. To that end, the scientific community and the businesses involved in AI are constantly investing in developing and implementing methods and technologies to enable high-performance on-device DNN inference.

Quantization is one of the most applied strategies to infer DNNs on edge devices. This is mainly due to its simplicity, straightforward implementation, and applicability to a wide range of scenarios, compared to hardware-specific solutions (e.g., Google Coral Edge TPU\footnote{\url{https://coral.ai/products/accelerator}}). In fact, quantization decreases DNN's memory requirements by lowering the precision of the values representing weights and activation from a 32-bit floating-point format to a much smaller one. With the industry's rising demand for quantization, the efficacy of the assessment of quantized DNNs is becoming an essential concern. To the best of our knowledge, no studies proposed dedicated techniques to assess the efficacy of these quantized  models. Instead, the classical approaches that are commonly used to assess vanilla (full-size) DNN models, are also adopted to assess their quantized versions.

Classical assessment methods judge a given DNN based on its accuracy on test data. When applied to a quantized DNN, they also judge its ability to preserve the performance of its original version. Even though this test data-driven assessment method has proved its effectiveness in evaluating DNNs, it might not be the best option to judge the impact of quantization on these models. In fact, test data are a set of labeled samples that give an insight on the nature, the features, and the properties of the processed data. In many domains, these samples are very hard to collect as they require expensive, time-consuming, and expert-depending procedures. This leads to smaller test datasets where the provided labeled samples might not properly represent the immensity of the input space. Besides, test data might not reflect every possible scenario that the model can encounter. In the same sense, they may not be able to trigger every possible divergent behavior between the DNN model and its quantized version. Hence, if we only rely on these traditional assessment methods, we may be led astray as two DNN versions of different arithmetic precision might show almost the same level of performance on test data, but trigger widely divergent behaviors when applied in real-world settings. 

In this paper, we address these challenges by proposing DiverGet, a customizable search-based software testing framework dedicated to quantization assessment. DiverGet aims to evaluate the real impact of quantization methods by detecting difference-inducing inputs that produce behavioral disagreements between an original DNN and its quantized version. To do so, DiverGet (i) extends the search space using semantically preserving metamorphic relations, then (ii) explores the search space following population-based metaheuristics, and (iii) uses diverse and complementary fitness functions to guide the search process and generate quality test data.

To demonstrate the effectiveness of our proposed DiverGet, we evaluate it on Hyperspectral Images (HSIs). These cubes of data are made of hundreds of bands that collect, each at a narrow interval of contiguous wavelengths, the energy emitted and/or reflected by the objects of the captured scene. Each band offers spatial information, and the stacked bands give us spectral information. This very rich information is extremely vulnerable to distortions caused by external factors. Besides, these rarely labeled data are frequently used in quantization-related applications.

Results using state-of-the-art HSI DNNs, namely the Hybrid Spectral Convolutional Neural Network (HybridSN)~\citep{roy_hybridsn:_2020} and the Spectral–Spatial Residual Network (SSRN)~\citep{zhong_spectralspatial_2018}, show that DiverGet succeeds in generating meaningful test inputs that induce behavioral disagreements among DNN versions. DiverGet also succeeds in outperforming the state-of-the-art testing framework, DiffChaser~\citep{xie_diffchaser:_2019}, in detecting divergences induced by quantization.

The contributions of the paper are summarized in the following:
\begin{itemize}
  \item We propose a customizable quantization assessment approach that combines metamorphic relations and population-based metaheuristics.
  \item  We evaluate the effectiveness of our approach using HSIs, since remote sensing is a growing and quantization-demanding domain. The results emphasize the importance of metamorphic relations in enriching the search space, and of metaheuristics in steering the search to prominent regions as opposed to random sampling.
  \item We conduct an advanced analysis to evaluate DiverGet's performance across numerous models, datasets, quantization approaches, fitness functions, and metaheuristic algorithms.
  \item We further demonstrate that DiverGet outperforms the state-of-the-art testing framework, DiffChaser~\citep{xie_diffchaser:_2019},  in detecting divergences induced by quantization.
\end{itemize}

The remainder of this paper is structured as follows: Section \ref{sec:bckgnd} introduces the fundamental concepts that will be used throughout our work. Section \ref{sec:RW} discusses related work. Section \ref{sec:meth} introduces DiverGet and describes its design process. Section \ref{sec:exp} reports the evaluation outcomes. Section \ref{sec:threats} analyzes the threats to the validity of our proposed DiverGet. Finally, Section \ref{sec:cl} concludes the paper.

\section{Background}
\label{sec:bckgnd}
In the following, we briefly describe the essential concepts of DNN quantization and search-based software testing.
\subsection{Quantization}
\label{sec:quantization}
Quantization is one of the most used compression methods to host practical-sized DNNs on edge devices. It converts the involved data tensors to have low-precision arithmetic types with the aim of downsizing the model footprint. For instance, a variant of quantization consists in decreasing the precision of weights and activations from single precision floating-point format (32 bits) to half precision floating-point format (16 bits) or integer format (8 bits)~\citep{guo_survey_2018}.

Among the benefits of the DNN's footprint reduction, we find faster computation, lower memory usage, as well as lower power consumption~\citep{krishnamoorthi_quantizing_2018}. Nonetheless, the resulting compression often comes with a degradation in the predictive performance of the quantized DNN. To assess this degradation, practitioners often gather as much labeled test data as possible and run it through the quantized DNN to quantify the drop in accuracy.

Recent quantization techniques perform well in keeping the accuracy of quantized DNNs equally matching their full-precision versions. In the following, we describe the most recent quantization methods~\citep{gholami2021survey, wu_integer_2020} that have been leveraged in our empirical evaluation:

\begin{itemize}
    \item Post Training Quantization (PTQ) compresses the DNNs by reducing the precision of either their weights or both their weights and activations, from 32-bit floating-point format to a lower format without requiring to retrain the model. It is a very simple approach and it permits quantization with little to no data.
    \item Quantization Aware Training (QAT) performs quantization during training by adding quantization operators to the DNN, to simulate the drop in precision in the forward propagation pass (the backward pass remains unchanged). This trick will add up the quantization error to the total loss and will make the optimizer reduce it appropriately. Therefore, QAT is known to provide better accuracies than PTQ, though it requires available training data.
\end{itemize}

\subsection{Search-based Software Testing (SBST)}
\label{SBST}
Search-Based Software Testing (SBST) dates from 1976 and back then, there was little interest in it. However, in the last decade, SBST has been used to solve a wide range of testing problems~\citep{mcminn_search-based_2011}.

SBST formulates the target test criteria as a fitness function that compares and contrasts candidate solutions from the search space in terms of their adequacy. To resolve the formulated optimization problem, SBST techniques leverage metaheuristic search algorithms, as gradient-free optimizers. These latter require few to no assumptions on the properties of both the objective function and the inputs search space. However, they do not provide any guarantee of finding an optimal solution. Their applicability in input test generation is suitable as these problems frequently encounter competing constraints and require near optimal solutions, and these metaheuristics seek solutions for combinatorial problems at a reasonable computational cost.

In line with the recent results on SBST applied for quantization assessment~\citep{braiek_deepevolution:_2019,xie_diffchaser:_2019}, we opt for nature-inspired population-based metaheuristics, such as Particle Swarm Optimization (PSO)~\citep{eberhart_new_1995} and Genetic Algorithm (GA)~\citep{mitchell_introduction_2001}. These latter possess intrinsically complex routines and non-determinism that make them high potential candidate for spotting vulnerable regions in the large, multi-dimensional input space of the Deep Learning (DL) models. 


\section{Related Work}
\label{sec:RW}
Over the last few years, researchers have adapted concepts and methods from software testing to invent advanced DL testing approaches~\citep{braiek2020testing}. One common denominator of these new testing approaches is their focus on generating synthetic test data with high ability to reveal erroneous behaviors of the DL models. Their test data generators consist in solving a maximization problem, i.e., finding the inputs that enhance the test adequacy criterion, relying on gradient-based optimizers~\citep{pei_deepxplore:_2017, ma_deepgauge:_2018}, gradient-free optimizers (i.e, metaheuristics)~\citep{braiek_deepevolution:_2019,xie_diffchaser:_2019}, and greedy search algorithms~\citep{xie_deephunter:_2018,odena_tensorfuzz:_2018,tian_deeptest:_2018}.

Although adversarial attacks~\citep{biggio2018wild} operate similarly by generating maliciously-crafted inputs, they maximize the probabilities of other labels or the closest one against the correct label's probability, aiming to alter the prediction of the DNN to yield wrong label. On the other hand, DL testing approaches adapt diverse test adequacy criteria, inspired from traditional software testing domain. Pei et al.~\citep{pei_deepxplore:_2017} proposed Neuron Coverage (NC), the ratio of newly-activated neurons by test inputs, which is inspired by conventional code coverage. Next, Ma et al.~\citep{ma_deepgauge:_2018} generalized NC by defining multi-granularity testing criteria, i.e., neuron-level and layer-level coverage. For instance, they refine NC to target (i) the major neuronal behaviors using K-multisection Neuron Coverage (KMNC), the ratio of the covered k-multisections of neurons, and (ii) minor corner-case neuronal behaviors using Neuron Boundary Coverage (NBC), the ratio of the covered boundary region of neurons.

While most of the DL testing approaches target the search of inconsistencies in the best-fitted DNN, a few of them consider the detection of disagreements between multiple DNNs with different arithmetic precision. TensorFuzz~\citep{odena_tensorfuzz:_2018} was the first method leveraging a coverage-guided fuzzing test generator to expose difference-inducing inputs between the best-fitted DNN and its quantized counterpart. DeepHunter~\citep{xie_deephunter:_2018} follows the same strategy of coverage-guided fuzzing, but enriches the data perturbations from only noise addition to $8$ different image-based transformations, including pixel-value perturbations and geometric transformations. DeepEvolution~\citep{braiek_deepevolution:_2019} builds on top of the last work by preserving the diversity of input data transformations. Additionally, it optimizes the generation of test inputs towards most prominent regions using metaheuristic-based search approach, and was instantiated by $9$ different nature-inspired population-based metaheuristic algorithms. However, the main limitation of all these early-released research works is their empirical evaluation on simple quantized models that are usually hand-crafted and weights-only.

Recently, DiffChaser~\citep{xie_diffchaser:_2019} challenged the state-of-the-art quantization methods supported by mainstream DL libraries, including Tensorflow Lite and CoreML. Its core approach relies on search-based generation of distorted inputs using GA to perform untargeted/targeted attacks to expose behavioral divergences amongst different subjects of low/high arithmetic precision DNNs. Despite significant advances over previous approaches that were generic and less effective, DiffChaser focused on the simple detection of the unavoidable existence of divergence between two DNN versions with different arithmetic precision. None of these mentioned studies assessed the quantization-induced divergence rate (i.e., the ratio of disagreement behaviors with respect to the total number of passed tests) and compared the rates obtained by different quantization techniques.

DiffChaser, the most recent DL testing framework for quantization degradation, will serve as a baseline for our empirical evaluation on the state-of-the-art HSIs DNN classifiers.


\section{Methodology}
\label{sec:meth}
In this section, we introduce our proposed approach for DNN quantization assessment and describe its principal aspects, namely the domain-specific metamorphic relations and the search-based data transformation.   
\subsection{Quantization Assessment via Systematic Detection of Behavioral Divergences}


DiverGet’s protocol for quantization assessment is designed to systematically search and expose behavioral divergences by showing that a quantized DNN behaves very differently on metamorphically-transformed inputs. DiverGet probes the quantized model’s erroneous behaviors w.r.t. its full-precision counterpart on practically-relevant data distortions. The main point is that our crafted synthetic inputs induce differences between the DNNs' inference behaviors, not simply conventional adversarial attacks that provoke mis-predictions by the quantized version. This variation distinguishes difference-inducing test inputs from the more familiar fault-inducing adversarial inputs. This strongly implies that techniques generating random noisy data with no true label, maliciously-perturbed inputs with ambiguous semantics, or highly-stretched data risking out-domain distribution, are insignificant for difference-inducing test exploration. The first reason is that even state-of-the-art models still suffer from these adversarial search techniques mainly because of their over-parameterization and high sensitivity to certain well-crafted changes. Thus, the resulting disagreement would not provide novel insights on the preserved robustness following the quantization since the original model would often predict a wrong output. The second reason is that it is less interesting if the two versions of the DNN disagree on semantically-ambiguous and practically-irrelevant synthetic test inputs. In fact, the reduction of the arithmetic precision in the quantized DNN would naturally induce numerical calculation differences reaching the final output, in certain subspaces of the input data.

In response to that, we propose to elaborate domain-specific metamorphic relations, that we later introduce in Section \ref{MR_section}. This allows realizing more task-oriented performance comparisons of the two DNN versions against input distortions that arise naturally from certain condition changes in real-world deployment settings.

The systematic aspect of our approach is established by the variation in the distortion type and its level of intensity within the permitted range. However, given the high-dimensionality of this derived synthetic input space, the trade-off between the cost of the assessment and the coverage of behavioral divergences should be well balanced and configurable according to the safety-critical aspect of the application or the user’s preferences.

On one side, we adopt a search-based approach (please refer to Section \ref{SBST_section}) that leverages metaheuristics to optimally drive the data transformations towards the production of difference-inducing inputs. We design two alternative fitness functions that would serve as different search objectives, depending on whether fast reaching the divergences or diversifying the triggered behaviors is the priority. On the other side, we choose population-based metaheuristics as the principal type of search algorithm in our assessment because of their exploration capabilities. Their parameters regarding the size of population and the number of iterations would be chosen in advance to control the amount of distortions generated by every session.

Furthermore, our protocol supports two modes, either sampling one original datapoint or a batch of datapoints, to apply on them the generated distortions. The batch mode reduces the cost of test generations as low as we increase the size of the batch, but the search for interesting distortions on a single original datapoint can be more focused, and hence, more effective.

Last, the revealed set of behavioral divergences between the original and the quantized models can be an additional quality criterion to compare between different quantization techniques, and can influence the decision-making process on the real-world deployment of the quantized model at the edge.

\begin{figure}[!h]
  \centering
  \includegraphics[width = \textwidth]{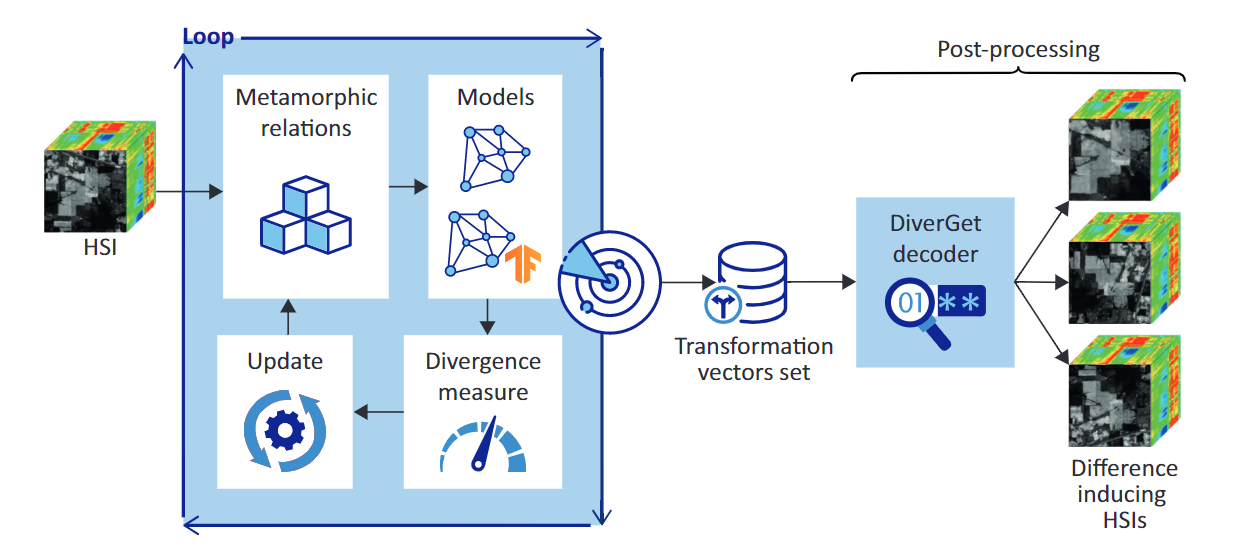}
  \caption{The workflow of DiverGet: starting from a given HSI (or a batch of HSIs), our proposed framework creates a set of transformations vectors. These latter can be used, in a post-processing step, to generate difference-inducing HSIs.}
  \label{fig:workflow}
\end{figure}

Figure \ref{fig:workflow} shows the overview of the systematic process of the proposed quantization assessment approach. 

\subsection{Search for Transformations instead of Data}
\label{data_trf}
Due to the high dimensionality of HSIs, we chose to change the search space by searching for data transformations, i.e., vectors encoding the steps of transforming an original HSI into a distorted HSI, rather than searching directly for HSIs. The transition from highly-dimensioned data to lower-dimensioned transformation simplifies the search problem by narrowing the search space and simplifying validation checks. While we cannot define the search space of all valid neighbors of an original image, we can constrain the parameters of the transformations with the aim of systematically promoting valid transformed inputs as results. 

The transformation vectors are the backbone of DiverGet’s workflow and can efficiently and easily help produce the ultimately needed distorted HSIs. The vectors that result in Difference-Inducing Inputs (DIIs) will be finally stored by DiverGet. These vectors provide all the necessary information (i.e., metadata) to generate DIIs in a post-processing stage where the user can easily utilize DiverGet’s decoder, as illustrated by Figure \ref{fig:workflow}. Generating and exploiting DIIs might be a beneficial and rapid approach for quantized model repair in comparison to repeating the training from scratch. In fact, current research has begun to investigate several strategies that leverage DIIs to fix the divergence between the two DNN versions~\citep{hu2022characterizing} (i.e., original and quantized versions).

The benefits of the transformation vectors are not limited to the saving of the storage space and the reconstruction of synthetic inputs. In fact, the valuable information encoded in these vectors may be exploited to gain insights into the causes of divergence between the model versions. In particular, statistics on these vectors can yield further explanations for the divergence pattern. For instance, a simple study of these vectors may indicate which transformations result in more divergences, or what correlations exist between them. These findings might be extremely useful to developers, however without transformation vectors, they would be difficult to pinpoint.

DiverGet’s purpose is to revisit the quantization assessment by adding the evaluation of the robustness of the quantized version against naturally-occurring input distortions, with respect to its original counterpart. Thus, tracking all the revealed DIIs is essential to estimate the overall divergence rate. This latter can be considered as a novel robustness measure that is complementary to other conventional metrics like the success rate that detects the existence of divergence between two DNN versions with different arithmetic precision. Searching for all possible DIIs may raise some scalability concerns. However, the continuous test of label differences (i.e., checking if the MR was violated) has a very low overhead since these labels were previously calculated to estimate the fitness values. The time complexity of this operation will be O(1) and thus does not hinder the scalability of DiverGet. Regarding the scalability of the storage, our tracking module follows a periodic offloading strategy to avoid high memory usage. In fact, we save periodically the difference-inducing transformations into the hard drive after reaching a pre-fixed size. In our tests, we set the size to 10 Mb in order to balance between the I/O operations and the memory consumption.

\subsection{Domain-specific Metamorphic Relations}
\label{MR_section}

Given a labeled dataset, $\mathcal{D} = {(X_i, y_i)\quad \forall i\in\{1\; ..\;N=|\mathcal{D}|\}}$, a model $m_o$ and its quantized version $m_q$, we design a space of semantic-preserving Metamorphic Relations (MRs), $\mathcal{R}$. Each MR, $r\in \mathcal{R}$, would map a compound HSI distortion, $T_r\in\mathcal{T}$, to the identity of the expected predictions, formulated as $r: (T_r(., \theta_r), \mathbbm{1})$, where $\theta_r$ are the parameter values of $T_r$ within their valid set, $\Theta$. Thus, the follow-up test consists of asserting that:
\begin{equation}
\centering
(m_o(T_r(X_i)) = m_q(T_r(X_i))) \wedge (m_o(X_i) = y_i),\quad \forall i\in \{1\; ..\;N\}
\end{equation}

Hence, a failed test would yield a synthetic input, $\hat{X}_i = T_r(X_i)$, that violates a MR, $r$, and will be considered as a DII.\\

MRs have been extensively leveraged to uncover anomalies of different generations of software systems. They define a relationship between input data transformations and their corresponding expected outputs. The strength of this simple concept is that well-designed MRs help automating the large-scale generation of test cases without human intervention.

In the following, we describe the steps and principles adopted to construct MRs that relate semantically-preserving HSI transformations to the identity relationship between the expected outputs. It is worth mentioning that the design of HSI transformations and their configurations were performed in collaboration with domain experts.

\begin{enumerate}[label=(\roman*)]
\item Codifying naturally-occurring distortions as probabilistic data perturbations:

In contrast to traditional color images made of three bands, remote sensing HSIs are large cubes composed of hundreds of spectral bands. Each of the latter records, at a narrow interval of contiguous wavelengths, the radiations reflected and/or emitted by the objects of the captured scene. These images offer extremely rich spatial and spectral information. Nevertheless, they are vulnerable to atmospheric changes and acquisition system's defects that often induce partial information losses or alterations, in a chaotic or a regular way.

Through collaboration with two HSI experts who have been working on remote sensing image processing since 1994 and 2015~\citep{zaatour2020unsupervised,bouzidi2019parallel}, respectively, we implemented distortions that mimic naturally-occurring alterations~\citep{agili_revue_2014} caused by instabilities or defects in acquisition systems, or by atmospheric influences. Domain experts~\citep{agili_revue_2014} project these anticipated degradations of information, affecting HSIs, into groups, including radiometric perturbations that alter the luminance values of the pixels, and geometric transformations that change the geometric structure of the image.

For classification requirements, HSIs are cropped into multiple overlapped 3D patches constituting the data inputs for DNNs. Numerically, a 3D patch is a tensor, $\textbf{X} = (x_{i,j,k}) \in \mathbb{R}^{d_{1} \times d_{2} \times d_{3}}$, with 2 spatial dimensions, $d_{1}$ and $d_{2}$, and one spectral dimension, $d_{3}$. We denote by $\textbf{X}^{dis} = (x_{i,j,k}) \in \mathbb{R}^{d_{1} \times d_{2} \times d_{3}}$, the distorted 3D patch.

Table \ref{tab:table_1} summarizes the details of the distortions we implemented and used through this paper. In this table, distortions are applied to randomly picked pixels, $x_{(i', j', k')}$. Furthermore, we emphasize that all the random values of any distortion parameter (e.g., mean of Gaussian noise, rotation angle), as well as all the arbitrary selections of targeted components (e.g., region, line, pixels) would be either uniformly sampled or optimally set by the optimizer, while always remaining within the possible range of values or set of choices.
\end{enumerate}

\begin{table}[!h]
\caption{Presentation of the used HSI distortions}
\label{tab:table_1}
\rowcolors{2}{blue!15}{}
\renewcommand{\arraystretch}{1.5}
\begin{tabular}{p{0.15\textwidth}p{0.11\textwidth}p{0.63\textwidth}}
\rowcolor{Blue}
\multicolumn{1}{c}{\textcolor{white}{Distortion}} & \multicolumn{1}{c}{\textcolor{white}{Variants}} & \multicolumn{1}{c}{\textcolor{white}{Short Description}} \\
Continuous Drop out & Line, Column, or Region & Replace the intensity values of \textit{all} the pixels belonging to a random component (e.g., line, column) by either the maximum or minimum intensity value within the 3D patch. \\ 
Discontinuous Drop out & Line, Column, or Region & Replace the intensity values of \textit{some} of the pixels belonging to a random component (e.g., line, column) by either the maximum or minimum intensity value within the 3D patch. \\ 
Stripping & Line or Column & Update the pixels belonging to a random component, $x_{i',j',k'}$, using the formula $\frac{\sigma _{d}}{\sigma_{o}}.(x_{i',j',k'}-m_{o}+\frac{\sigma_{o}}{\sigma_{d}} m_{d})$, where ($m_{o}$, $\sigma_{o}$) and  ($m_{d}$, $\sigma_{d}$) are the pair (mean, standard deviation) of the original and distorted patches, respectively.\\ 
Spectral band loss & NA & Replace an arbitrary subset of band values with the mean of their successor and predecessor.
\\ 
Salt and pepper noise & NA & Scatter arbitrary bright pixels (salt) or dead pixels (pepper) using, respectively, the maximum or the minimum value in the 3D patch. \\ 
Gaussian noise & Spectral or Spatial & Add a random noise $\phi$ that follows a Gaussian distribution $(\phi \sim N(\mu, \sigma ))$, to either the spectral or spatial information of arbitrary pixels.\\ 
Rotation & NA & Rotate the patch around its central pixel, by a randomly-picked angle $\alpha$.
\\ 
Zoom & In or Out & Perform in or out zooming centered on the central pixel of the patch according to an arbitrary zoom factor.
\\
\end{tabular}
\end{table}

\begin{enumerate}[label=(\roman*)]
\setcounter{enumi}{1}
\item Pre-define and set up the appropriate range of distortions:

The range of possible values for each transformation's parameters would decide the resulting levels of distortions on the original data. In fact, a wide range would result in high loss of information that have high chances to produce meaningless inputs that can be filtered upfront, i.e., images that are hard to occur in real-world settings. Inversely, a narrow range would be too conservative with meaningful inputs that are very close to their original parent and have low chances to trigger uncovered behaviors of the DNN. Hence, it is important, in a pre-processing phase, to set up the appropriate range of distortions that balance between enhancing the diversity and preserving the semantic identity of the synthetically-crafted images.

To do so, we use the Peak Signal-to-Noise Ratio (PSNR)~\citep{shi2017image}, a widely-used metric in the HSI field, to assess how much information loss occurs after the distortion. Thereby, we adjust the range of parameters of each codified distortion to reach or be close to their appropriate ranges. We start by the full range of possible values, then, we narrow down the range if the PSNR of the produced inputs are generally low. The significant degradation of the obtained PSNRs indicates that the distorted images are becoming out of the original data distribution.\\ \\

\item Ensure the validity/meaningfulness of the distorted inputs:

Despite the tuning of the ranges of the distortions' parameters, there is no guarantee that the synthetic images would be in-domain distribution and preserve the semantic identity of their genuine parents, especially when multiple distortions are applied in a compound transformation. 

To ensure the meaningfulness of the generated distorted inputs, we opted for: (i) a \emph{conservative strategy:} We adhered to expert guidelines throughout the conception of our synthetic data to ensure plausible real-world application. For instance, we constrain radiometric distortions to either be column- or row-wise due to the type of acquisition system, which collects data transversely, either by columns or by lines.

Furthermore, we set up a  (ii) follow-up post-distortion validity test where we reused the PSNR for the second time. If the PSNR of distorted patches is less than 20dB (experts' recommendation), they are discarded as invalid inputs. In fact, PSNR is utilized in HSI compression to verify the image quality and ensure semantics are preserved~\citep{yang2007optimal, dua2020comprehensive}. Thus, we apply the same validity check for our synthetic test inputs, using the default threshold of $20$ dB as stated in Section 5 of~\citep{thomos2005optimized}.
\end{enumerate}

\subsection{Search-based Approach for Data Transformation}
\label{SBST_section}
Once we have the domain-specific metamorphic relations, we aim to systematically search for the pairs of (original datapoint, crafted distortion) that enhance the diversity of the divergence-oriented test cases and also increase their ability to reveal disagreements between the best-fitted DNN and its corresponding quantized variants. Below, we detail the development steps that we follow to build the underlying search-based approach. 

\subsubsection{Definition of the Search Space}
\begin{figure}[!h]
  \centering
  \includegraphics[scale=0.5]{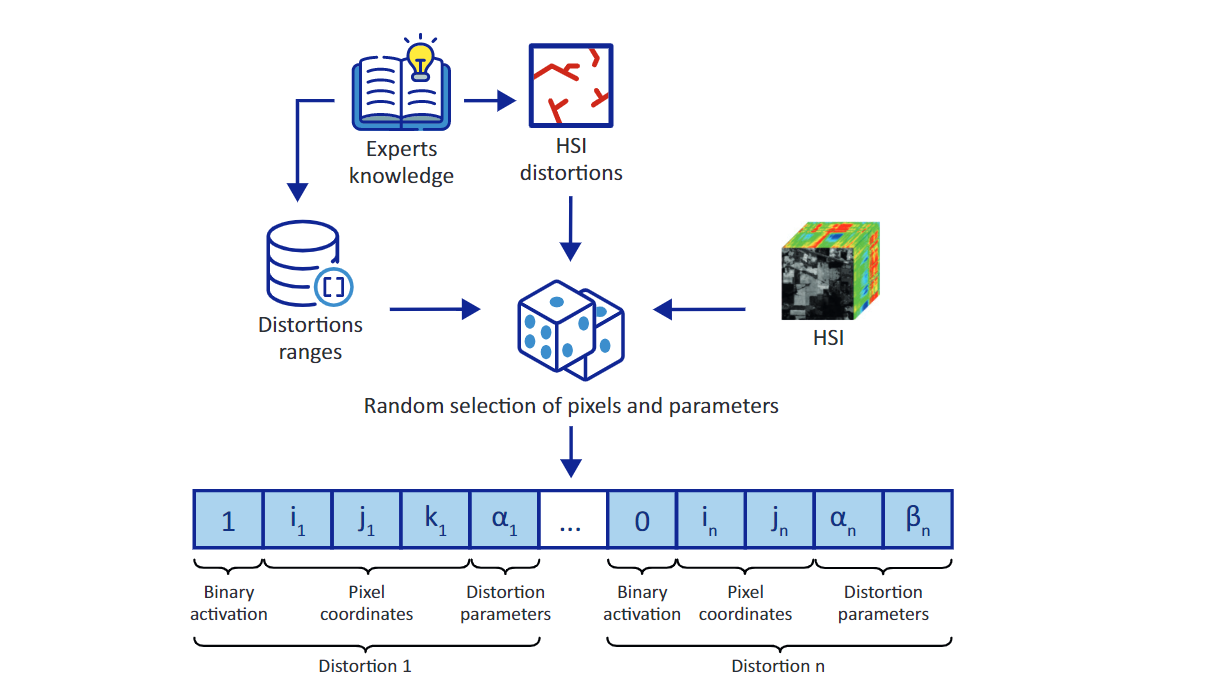}
  \caption{Formatting all distortions into a transformation vector}
  \label{fig:systematic_approach}
\end{figure}

Given the complexity of our codified input distortions, it is difficult to formulate the space of possible distorted versions from an original datapoint, contrary to other SBST for quantization assessment that restrict the data transformation to pixel-value noise addition (i.e., the search space for an input $x$ would be $[x-\delta, x+\delta]$, where $\delta$ is the maximum allowed pixel-value deviation). Thus, we choose to define the space of possible data transformations, then, the synthetic test inputs would be the result of the application of the crafted data transformations, and an original input data. To do that, we stack (i) the parameters in relation to all the supported data transformations, (ii) the binary variables that serve as a trigger to signal whether or not a specific distortion is active, and (iii) the integer variables that represent distorted pixel coordinates. For example, to encode rotation we first conserve the activation binary switch, followed by the coordinates of the central pixel, and finally the rotation angle. Therefore, the search space of data transformations is a multi-dimensional vector space, $T\in\mathbb{R}^k$, where each component/axis represents either a parameter value, an activation value, or a coordinate value. (see Figure \ref{fig:systematic_approach}).

\subsubsection{Design of the Fitness Functions}
As our main objective is to expose high density and diversity of difference-inducing test inputs, we should design a fitness function, $f_X$ that measures how much the synthetic input $\hat{x}$ induces behavioral divergences between the original DNN and its quantized version. Nevertheless, we define the data transformation space $T$ to be our search space. The optimizer would use a fitness function, $f_T$, to compare and contrast the fault-revealing ability of data transformations, $t\in T$, and the connection between both fitness functions, $f_X$ and $f_T$, would simply be $f_X(\hat{x}) = f_T(t)$, where $\hat{x} = t(x)$.

In the following, we discuss two specialized fitness functions that have been designed to promote the fault-detection capability and the diversity of the generated test inputs.

\begin{enumerate}[label=(\roman*)]
\item Divergence-based Fitness Function: 

Given two DNNs, the original model, $m_{o}$, and its quantized version, $m_{q}$, we denote $\textbf{\textit{s}}_{o}$ and $\textbf{\textit{s}}_{q}$ their respective softmax activation functions. Softmax is the last layer (output) activation that transforms the logit scores into probability distribution, where its component is the probability of a class label membership. The predicted label for the input is the one with the highest probability. Given a $c$-classification problem and a vector of logits, $\textbf{\textit{l}} = (l_{1}, ..., l_{c}) \in \mathbb{R}^{c}$, the softmax, $\textbf{\textit{s}} = (s_{1}, ..., s_{c}) \in \mathbb{R}^{c}$ is computed as follows:
\begin{equation}
  s_{i} = \sigma (l_{i}) = \frac{e^{\,l_{i}}}{\sum_{j=1}^{c} e^{\,l_{j}}} \quad for\quad i= 1, ..., c
\end{equation}
We aim to design a fitness function that estimates the divergence between the two probability of class membership distributions obtained by the original and quantized DNNs. Indeed, maximization of this divergence measure would increase the chances of disagreement between the two models caused by a mismatch of their two most-probable labels for the same input. There are two common divergence measures to compare different probability distributions, Kullback–Leibler divergence (KLD)~\citep{Kullback1987LETTERTT} and Jensen–Shannon divergence (JSD)~\citep{cover_elements_1991}. In our divergence-based fitness function, we opt for JSD, denoted $J$. Considering two probability distributions, $Q$ and $R$, defined on the same probability space $\chi$ $(|\chi|=c)$, $J$ can be formulated as follows:\\
\begin{equation}
  J(Q||R) = \frac{1}{2} (D(Q||M) + D(R||M)) 
\end{equation}
where $D(Q||M) = \sum_{i=1}^{c} Q(i) \ln(\frac{Q(i)}{M(i)})$ and  $M = \frac{1}{2}(Q+R)$. 

The main reason we chose JSD over KLD is that the former is symmetric, $J(Q||R) = J(R||Q)$, and bounded, $0\leq J(Q||R)\leq 1$. Therefore, we define, below, our divergence-based fitness function of the synthetically-produced inputs.
\begin{equation}
  f_{X}^{div}(\hat{x}) = J(\textbf{\textit{s}}_{o}(\hat{x}), \textbf{\textit{s}}_{q}(\hat{x}))
\end{equation}

This divergence-based fitness function can be considered as grey-box testing criterion because it has access only to the last layer softmax activations. Despite it targets directly the enhancement of the distance between the two models' predictions, it risks the phenomenon of mode collapse, where the optimizer falls on a local maximum (i.e., the transformed inputs trigger high or even the maximum divergence). Hence, all the generated inputs would be very similar or even identical. To mitigate this probable issue, we propose the following alternative fitness function.

\item Coverage-based Fitness Function:

We aim to design a white-box testing criterion that estimates differences in the signature of neurons activations obtained from both tested versions of the DNN. This helps compute at fine-grained level the behavioral divergences caused by the test input and ensures that the optimizer would continuously search for the synthetic inputs, inducing higher differences at all the layers' activations.

To do that, we rely on K-multisection Neuron Coverage (KMNC) that is inspired by the Neuron Coverage. The adoption of KMNC was motivated by the diversity enhancement in test input generation achieved by neuron coverage. In fact, diversifying inputs helps prevent mode collapse. Pei et al.~\citep{pei_deepxplore:_2017} and Tian et al.~\citep{tian_deeptest:_2018} have demonstrated that increasing the neuron coverage is highly related to enhancing the diversity of the generated test cases. In the same vein, Yu et al.~\citep{yu2019test4deep} have shown that coverage-guided test generation produces inputs with increasing L1 distances from their original seeds. Indeed, KMNC does not consider each neuron as active ($1$) or inactive ($0$) depending on a predefined threshold like NC, but instead, it refines this over-approximate binarization of the continuous neuron output into a multi-section discretization.

Let $N = \{n_1,n_2, . . .,n_M\}$ be the set of neurons of a given DNN and $I = \{i_1,i_2, . . .,i_M\}, \; i_m=[low_m, high_m]\quad \forall m\in\{1\; ..\;M\}$, be the set of neurons activations' intervals observed during the training, KMNC divides all the intervals into $k$ equal sections that should be covered by the test data, as an indication that the major behaviors of the trained DNN are triggered at least once during the testing.

As we are interested in defining a signature of neurons activations triggered by an input $x$, we exploit KMNC as detailed below.

Let $S_{i}^{n_m}$ be the set of activations from the neuron $n_m$ in the $i$-th section for $1 \leq i \leq  k$. Therefore, $\phi(\textbf{x},n_m) \in S_{i}^{n_m}$ indicates if the $i$-th section of the neuron $n$ is covered by the input \textbf{x}. Then, we define the signature $S^x$ as the subset of all the neurons' sections covered by $x$, as follows:
\begin{equation}
S^{x} = \{ S_{i}^{n_m}| \phi (\textbf{x},n) \in S_{i}^{n}\}, \quad \forall m\in\{1\; ..\;M\}\}
\end{equation}

Given an original and quantized DNN models, we can infer their corresponding signatures of neurons' activations for the same input $x$, respectively, denoted as $S_o^{x}$ and $S_q^{x}$. As each of them represents a set of covered sections, we can use the Jaccard similarity coefficient~\citep{jaccard_distribution_1912}, $J_{sc}$, which is a well-known similarity metric for comparison between two sets.
\begin{equation}
J_{sc}(S_o^{x} ,\, S_q^{x} ) = \frac{\left | S_o^{x}  \, \cap \, S_q^{x} \right |}{\left | S_o^{x}  \right | + \left | S_q^{x} \right | + \left | S_o^{x} \, \cap \, S_q^{x} \right |}
\end{equation}

Thereby, we design the coverage-based fitness function, $f_{X}^{cov}$, that assesses the dissimilarities in the neurons activations state obtained from the two DNN versions in order to encourage the generated test inputs intensifying these dissimilarities, and hence, expose hidden disagreements between their final predictions.

\begin{equation}
  f_{X}^{cov}(\hat{x}) = -J_{sc}(S_{o}^{\hat{x}}, S_{q}^{\hat{x}})
\end{equation}\\

The design of $f^{cov}$ is inspired by the feature matching~\citep{salimans2016improved} concept from Generative Adversarial Networks (GANs). GAN suffers from mode collapse when the generator over-exploits the mistakes previously made by the discriminator and fails to explore beyond them. To stabilize GAN training and decrease mode collapse, a new objective function design was proposed to steer the generator towards matching the expected values of the discriminator’s hidden layers, rather than matching only the discriminator's output.
\end{enumerate}

Although the second fitness function takes into account the internal neuron activities and estimates the internal behavioral divergences at fine-grained level, there is no guarantee that the generated test inputs, emphasizing these neuron-level differences between two under test models, would provoke mismatches of their predicted labels, as evidenced by former coverage-guided approaches~\citep{xie_deephunter:_2018, odena_tensorfuzz:_2018}. From this perspective, we assume that the first fitness function based on the divergence of softmax outputs can be more effective in driving the test input generation towards enlarging the gaps between the two probability of class membership distributions obtained by the original and quantized DNNs, until they predict two unlike most probable labels.

Moving on to the fitness evaluation of the generated data transformation, in our approach, we support two modes. The first is the single instance mode where only one original input $x^i$ is fixed in the loop, so that all the generated transformations $t^j\in T$ would be evaluated based on the fitness $f_T(t^j) = f_X(\hat{x}^{i,j})$, where $\hat{x}^{i,j} = t^j(x^i)$. The second is the batch mode that fixes a subset of $B$ original inputs, $X_b$, at once, so that any data transformation $t^j\in T$ would be evaluated based on the average fitness $\bar{f}_T(t^j) = \frac{\sum_{i=1}^{B}{f_X(\hat{x}^{i,j})}}{B}$, where $\hat{x}^{i,j} = t^j(x^i), \quad \forall i\in \{1\;..\;B\}$. For the sake of simplicity, we present the formulation of our above-mentioned fitness functions on the single instance mode.

It is worth highlighting that, in the scope of this work, we define two fitness functions. Nevertheless, DiverGet provides a plug-and-play design in which users may specify their preferred fitness functions. In reality, a variety of alternative criteria might be used to determine divergence. The work of~\citep{alzantot_genattack:_2019}, in which authors attempt to evaluate a model's robustness through adversarial attacks, might serve as an inspiration for the design of a new fitness function. The authors designed a straightforward function that assesses the model’s output score given to the target class label. In the same spirit, and with minor tweaks to binarize this fitness function, one may compute the difference between the log of the highest class probability predicted by the original model and the log of the one predicted by its quantized counterpart. Another fitness function design could be inspired by the prediction uncertainty introduced in DiffChaser’s paper~\citep{xie_deephunter:_2018}. The prediction uncertainty indicates that the DNN is uncertain whether to classify the input $x$ as $y$ or $y’$ because their predictive probabilities are very close. Following this idea, the new fitness function could compute the distance between the two highest class probabilities that are predicted the original model when having $x$ as input.

\subsubsection{Implementation of Metaheuristic-based Optimizers}
As introduced in Section \ref{SBST}, we apply SBST, using population-based metaheuristics, to drive optimally the transformation generation towards diverse prominent regions in the search space. Algorithm \ref{alg:algo1} explains the generic structure and steps of our approach to generate data transformations. Additionally, DiverGet supports divergence-based and coverage-based fitness functions as criteria for test adequacy, and it is also compatible with any population-based metaheuristic algorithm. The proposed algorithm starts with an initial random population of $p$ transformation vectors (Line $2$). The  synthetic  test  inputs  would  be  the  result  of  the  application of these data transformations (transformation vectors) and our original input data (Line $5$). Then, the algorithm computes the fitness values for all the population individuals, depending on the selected fitness function and the evaluation mode (Line $6$). Based on the obtained fitness values, the chosen metaheuristic algorithm applies its update routines on the population to derive the next generation of new candidates that are strong and likely better than their predecessors in terms of fitness (Line $13$). The enhancement of the fitness values would lead to synthetic inputs that enlarge further the behavioral deviations between the two tested DNNs either at the last layer's level or hidden layers' level, depending on the choice of the fitness function. Hence, we have high chances to find DIIs among the produced generations over the iterations. However, we pass validity test on the transformed inputs based on their resulting PSNR measures and only the valid ones (their PSNR values are more than $20$ dB) would be considered for the difference-inducing selection criterion (Lines $7-12$) (see Section \ref{MR_section} (iii)). Indeed, those lines are the major change with regards to the standard steps of population-based metaheuristics. This change alters the objective of the optimizer from searching for an optimal solution (i.e., candidate having the highest fitness value) to tracking all the revealed DIIs among the evolving valid candidates throughout the generations. Last, the generation loop for transformations on one or multiple inputs, depending on the activated fitness evaluation mode, is repeated until the maximum number of generations is reached or when the early stop condition (i.e., no changes on either fitness values or number of DIIs) is encountered (Line $14$). 

\renewcommand{\algorithmicrequire}{\textbf{Input:}}
\renewcommand{\algorithmicensure}{\textbf{Output:}}
\begin{algorithm}
\caption{Data Transformation Generation}
\label{alg:algo1}
\begin{algorithmic}[1]
\Require $x$: input/ batch of input, $m_{o}$: original DNN,$m_{q}$: quantized DNN, p: population size, maxiter: max iterations
\Ensure D: Set of difference-inducing transformation vectors
\State $D :=  \emptyset$;
\State initial random population $T$; \Comment{population of $p$ data transformation vectors}
\State iteration := 0;
\While{$iteration < maxiter$}
\State $\hat{X}$ := \textbf{GenerateHSIs($T$, $x$)}; \Comment{by applying every  $t \in T$ on $x$}
\State $F$ := \textbf{ComputeFitness($\hat{X}$)}; \Comment{fitness function for each $\hat{x} \in \hat{X}$}
\State $\hat{X}_{valid}$ := \textbf{CheckValidationConstraint($\hat{X}$, $x$)}; \Comment{$\hat{X}_{valid} \subset \hat{X}$}
\For{ $ \,  \hat{x}_{d} \in \hat{X}_{valid}$ }
    \If{ \,  $isDII(\hat{x}_{d}, m_{o}, m_{q})$ } \Comment{$\hat{x}_{d}$ breaks a MR($m_{o}(\hat{x}_{d}) \neq m_{q}(\hat{x}_{d})$)}
        \State D := $D \cup \{{t}_{d}\} $; \Comment{${t}_{d}$ is the vector leading to $\hat{x}_{d}$}
    \EndIf
\EndFor
\State $T$ := \textbf{UpdatePopulation($T$, $F$)};
\If{ $isEarlyStopping(D, F)$}
    \State break;
\EndIf
\State iteration := iteration + 1;
\EndWhile
\end{algorithmic}
\end{algorithm}

In line with the No Free Lunch Theorem (NFL)~\citep{ho2002simple}, we implement two concurrent nature-inspired population-based metaheuristics, PSO and GA, that have shown their effectiveness in high-dimensional search problem such crafting black-box adversarial attacks~\citep{mosli_they_2019, alzantot_genattack:_2019}. For each implemented metaheuristic, we tune its hyperparameters to appropriately set up its level of non-determinism and of balance between the intensification (i.e., exploitation of the best candidates found to concentrate the search on the prominent regions) and diversification (i.e., the exploration of non-visited regions to avoid missing potential interesting solutions)~\citep{joshi2020parameter}.

\section{Experiments}
\label{sec:exp}
We evaluate the effectiveness of DiverGet through the following research questions (RQ):
\begin{itemize}
    \item \textbf{RQ1:} How effective is DiverGet's main feature (i.e., the domain-specific metamorphic relations and the search-based data transformation) at finding difference-inducing inputs?
    \item \textbf{RQ2:} How would DiverGet perform on numerous models, datasets, quantization approaches, fitness functions, and metaheuristic algorithms?
    \item \textbf{RQ3:} How does DiverGet compare to state-of-the-art DiffChaser?
\end{itemize}

\subsection{Experimental Setup}
In this subsection, we detail the different elements of our experimental setup.
\begin{itemize}
    \item Datasets: We consider two popular and publicly-available\footnote{\url{http://www.ehu.eus/ccwintco/index.php/Hyperspectral_Remote_Sensing_Scenes}} hyperspectral datasets. The first is Pavia University (PU), a HSI made of $610 \times 340$ pixels with $103$ spectral bands in the wavelength range of $430$ to $860$ nm. Its ground truth shows $9$ urban land-cover classes. The second is Salinas (SA), a HSI made of $512 \times 217$ pixels and $204$ spectral bands in the wavelength range of $360$ to $2500$ nm. Its ground truth contains $16$ vegetation classes.
    \item DNNs: To perform a classification of land covers using a CNN, the HSI is cropped into multiple 3D patches, with a prefixed size, that would be used as model inputs. For a better diversity of the evaluated subjects, we studied state-of-the-art CNNs, namely SSRN~\citep{zhong_spectralspatial_2018} and HybridSN~\citep{roy_hybridsn:_2020}, that are quite popular, open-source, and previously tested on the PU and SA datasets.
    
    SSRN is an end-to-end spectral-spatial residual CNN. It relies on 3D convolutional layers as a basic element of its architecture. It consists of two spectral and two spatial residual blocks that learn discriminative features from spectral signatures and spatial contexts in the HSI.
    
    HybridSN is a spectral-spatial CNN that combines both 3D and 2D convolutional layers. It uses 3D convolutional layers to learn the joint spatial-spectral feature representation on latent spatial features that are extracted by upper-level 2D convolutional layers.
    
    Table \ref{tab:table_2} gives an overview of these CNNs.

\begin{table}[!h]
\centering
\caption{Details of DNNs and datasets used to evaluate DiverGet. The Overall Accuracies (OAs) are taken from~\citep{roy_hybridsn:_2020, zhong_spectralspatial_2018}}
\label{tab:table_2}
\rowcolors{1}{blue!15}{}
\renewcommand{\arraystretch}{1.5}
\arrayrulecolor{Blue}
\begin{tabular}{cccc}
\rowcolor{Blue}
\textcolor{white}{Dataset} & \textcolor{white}{DNN Model} & \textcolor{white}{\# 3D patches} & \textcolor{white}{OA (\%)} \\ 
 & SSRN & Train, Val: 4281; Test: 34214 & 99.79\\ 
\multirow{-2}{*}{\cellcolor{white}PU} & HybridSN & Train: 12832; Test: 29944 & 99.98 \\ 
\hline
\cellcolor{blue!15} & SSRN  & Train, Val: 5418; Test: 43293 & 99.98\\ 
\hhline{>{\arrayrulecolor{blue!15}\doublerulesepcolor{blue!15}}->{\arrayrulecolor{black}}~~~}
\rowcolor{blue!15}
\multirow{-2}{*}{\cellcolor{blue!15}SA} & HybridSN & Train: 16238; Test: 37891 & 100 \\
\end{tabular}
\end{table}

\item Quantization methods: For more credibility and reproducibility, we chose to use real-world quantization methods, implemented by state-of-the-practice DL tools. More precisely, we opted for Tensorflow Lite\footnote{\url{https://www.tensorflow.org/lite}}, Google's open-source DL library for on-device deployment. For each original DNN, i.e., implemented in 32-bit floating precision, we performed $2$ types of Tensorflow Lite's provided quantization techniques, namely the 8-bit post-training quantization and the 8-bit quantization-aware training.
\item Evaluation metrics:
    \begin{itemize}
    \item Metrics of models divergence: We use three metrics to evaluate the divergence between the model versions: (i) the number of difference-inducing inputs (\#DII), (ii) the divergence rate (DiR), i.e., the percentage of difference-inducing inputs discovered over all the generated inputs for each seed, and (iii) the success rate (SR), i.e., the percentage of original 3D patches where at least one disagreement is successfully revealed among their descendant synthetic inputs.
    
    These metrics try to assess the divergence from different angles. In fact, DiR (and consequently \#DII) tries to quantify the degree of divergence between the two model versions by reporting all possible detected DIIs. Whereas, SR tries to detect the existence of divergence between the two models.

\item Metric of input validation: As introduced in Section \ref{MR_section}, the validity of synthetic HSI is defined by their PSNR value compared to the original HSI, to verify the image quality and ensure that semantics are preserved. If the PSNR of distorted patches is less than 20dB (experts' recommendation based on previous work), they are discarded as invalid inputs. Therefore to inform about the quality of generated HSIs (transformations that lead to HSIs), we define the Validation Rate (VR) metric as the percentage of valid-generated transformations/HSIs over all generated transformations/HSIs, for each seed.

\item Metric of execution time: To evaluate the execution time of their framework, the previous work~\citep{xie_diffchaser:_2019} has focused on quantifying the time required to detect the first disagreement. We followed the same approach and computed for each original HSI, the time required to identify the first disagreement per patch (FDI). In addition, as indicated in the work of~\citep{xie_diffchaser:_2019}, we measured the time required to identify the first disagreement per patch only for successful cases (i.e., original HSIs that resulted in DIIs) (FDI*). This measure is applied per patch only for successful cases (i.e., HSIs that lead to DIIs) (FDI*). The FDI* measure allows reporting the time without penalization to the framework in the absence of disagreement on a particular patch.

\item Metric of statistical significance and effect size: We use statistical hypothesis testing and effect size measurements to assess the statistical significance of our results. As our data are unlikely to be normally distributed, we utilized non-parametric hypothesis testing. Additionally, a paired test was chosen because of our belief in the interdependence of the two distributions. Specifically, we performed the Wilcoxon Signed Rank test~\citep{wilcoxon_individual_1945} and Vargha-Delaney   12~\citep{vargha2000critique} effect size test to determine how much two groups differ from one another.

If two groups are statistically indistinguishable, $\hat{A}_{12} = 0.5$. $\hat{A}_{12} > 0.5$ indicates that, on average, the first group outperforms the second, $\hat{A}_{12} < 0.5$ signifies that the second group outperforms the first. The magnitude of the difference between the groups can be classified into four categories based on the scaled  $\hat{A}_{12}$~\citep{hess2004robust}: “negligible” ($| \hat{A}^{scaled}_{12} | < 0.147$), “small” ($0.147 \leq | \hat{A}^{scaled}_{12} | < 0.33$), “medium” ($0.33 \leq | \hat{A}^{scaled}_{12} | < 0.474$), and “large” ($| \hat{A}^{scaled}_{12} | \geq  0.474$).
\end{itemize}
    \item Software: We developed DiverGet in Python. It supports Tensorflow (version 2.4.1)~\citep{abadi_tensorflow:_2016} and Tensorflow Lite models. For the metaheuristics implementation, we adapted the open-source python libraries, pyswarm\footnote{\url{https://pythonhosted.org/pyswarm/}} and geneticalgorithm\footnote{\url{https://github.com/rmsolgi/geneticalgorithm}}, to meet our design of population-based searching algorithm specifications.
    \item Hardware: We run all experiences on B2S instances of Microsoft Azure Virtual Machines. Each of these has a 2-core 2.4 GHz Intel Xeon CPU and 4 GB of RAM, and runs on ubuntu 18.04.
\end{itemize}

\begin{figure}[!h]
  \centering
  \includegraphics[width = \columnwidth]{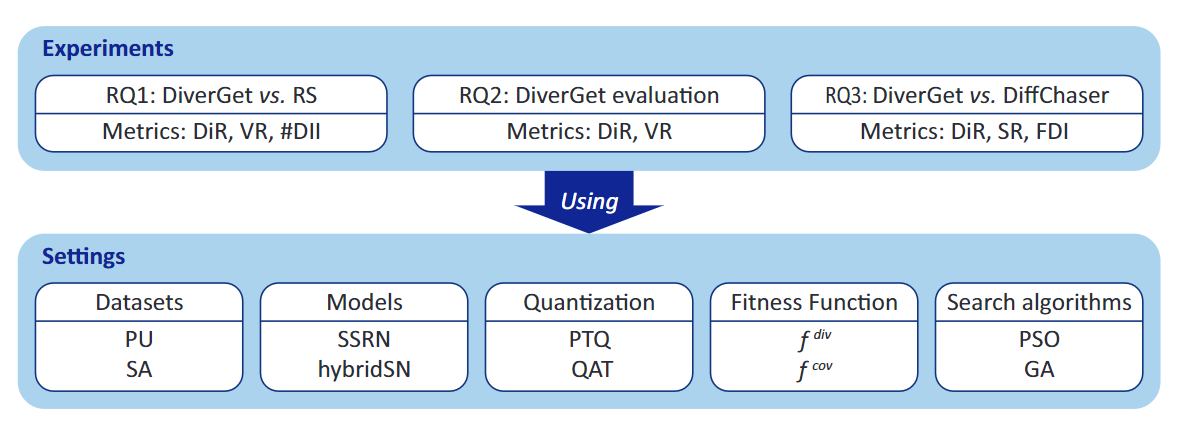}
  \caption{Overview of the experiment design and evaluation metrics.}
  \label{fig:fig3}
\end{figure}

The next sections describe the experiments that were carried out to answer our research questions (please consult Figure \ref{fig:fig3} for an overview of the experiment design).

\subsection{RQ1: The Effectiveness of DiverGet}

\textbf{Motivation:} Throughout its process, DiverGet (i) relies on domain-specific metamorphic relations to provide semantically preserving data and (ii) leverages various population-based metaheuristic algorithms to guide the search process and generate quality test data.

\textbf{Method:} For each dataset, we randomly select 10 seeds from a pool of 800 original inputs. All the original test inputs are sampled from the subset of test datasets, for which the original model correctly predicts their corresponding labels. For each model, we compare the original version to a quantized version using either PTQ or QAT. For the first experiment (i.e., evaluating the effectiveness of the proposed metamorphic distortions), we set a random sampler (RS) to generate synthetic data using our defined domain-specific metamorphic relations. We then compared the number of difference-inducing inputs (DII) found in the original test data and those generated by RS.

For the second experiment (i.e., evaluating the effectiveness of our searching strategy), we compared for each seed the RS equipped with our metamorphic distortions to DiverGet using various configurations. We compare the results of the second experiment using two metrics: i) the divergence rate (DiR), i.e., the percentage of difference-inducing inputs discovered over all the generated inputs for each seed, ii) the Validation Rate (VR), i.e., the percentage of valid-generated distortions (PSNR $> 20dB$) over all generated distortions, for each seed.

\textbf{Results:} Table \ref{tab:table_3} reports the number of DIIs revealed by the original test data and the synthetic inputs randomly sampled from the specified metamorphic relations. Tables \ref{tab:table_4} reports the median of the selected measurements (i.e., DiR, VR) of all used seeds when comparing DiverGet to RS. The values obtained by PSO and GA are averaged to represent the column ‘DiverGet’ in Table \ref{tab:table_4}.


\begin{table}[!h]
\centering
\caption{Comparison of the number of DII found in the original test data and by RS using all seeds}
\label{tab:table_3}
\rowcolors{1}{blue!15}{}
\renewcommand{\arraystretch}{1.5}
\arrayrulecolor{Blue}
\begin{tabular}{ccccc}
\rowcolor{Blue} \textcolor{white}{Model} & \textcolor{white}{Dataset} & \textcolor{white}{Quantization} & \begin{tabular}[c]{@{}c@{}}\textcolor{white}{\# DII - Original} \\ \textcolor{white}{Test Data}\end{tabular} & \begin{tabular}[c]{@{}c@{}}\textcolor{white}{\# DII - Random} \\ \textcolor{white}{sampling (RS)}\end{tabular} \\ 
 & {\cellcolor{white}} & PTQ & 136 & 1609  \\
 \cellcolor{white}& \multirow{-2}{*}{\cellcolor{white}PU} & QAT & 1 & 763  \\ 
 & {\cellcolor{blue!15}} & PTQ & 0 & 132  \\ 
 \multirow{-4}{*}{\cellcolor{white}SSRN}& \multirow{-2}{*}{\cellcolor{blue!15}SA} & QAT & 40 & 498  \\ 
 \hline
 \cellcolor{blue!15} & & PTQ & 0 & 133  \\
 & \multirow{-2}{*}{\cellcolor{white}PU} & QAT & 1 & 522  \\
\cellcolor{blue!15} & \cellcolor{blue!15} & PTQ & 0 & 110  \\ 
 \multirow{-4}{*}{\cellcolor{blue!15}hybridSN} & \multirow{-2}{*}{\cellcolor{blue!15}SA} & QAT & 10 & 506  \\
\end{tabular}
\end{table}

\textit{\textbf{Naturally-occurring synthetic inputs vs original test inputs: }} Table \ref{tab:table_3} shows that random samples of synthetic inputs expose more difference-inducing test cases than those of original inputs. Thus, the designed input distortions validate application-specific robustness requirements using corner-case test scenarios that cannot be assured directly from original test cases.

\begin{tcolorbox}[colback=blue!5,colframe=blue!40!black]
\textbf{Finding 1:} The designed domain-specific metamorphic relations expose uncovered behavioral divergences resulting from the quantization that original test data failed to shed light on.
\end{tcolorbox}


\begin{table}[!h]
\centering
\caption{Comparison of the median values of DiR and VR between RS and DiverGet}
\label{tab:table_4}
\rowcolors{2}{}{blue!15}
\renewcommand{\arraystretch}{1.5}
\arrayrulecolor{Blue}
\begin{tabular}{ccccccc}
\rowcolor{Blue}
& & & \multicolumn{2}{c}{\textcolor{white}{RS}} & \multicolumn{2}{c}{\textcolor{white}{DiverGet}} \\
\rowcolor{Blue}
\multirow{-2}{*}{\textcolor{white}{Model}} & \multirow{-2}{*}{\textcolor{white}{Dataset}} & \multirow{-2}{*}{\textcolor{white}{Quantization}} & \textcolor{white}{DiR} & \textcolor{white}{VR} & \textcolor{white}{DiR} & \textcolor{white}{VR} \\
& & PTQ & 1.07 & 3.75 & 24.05 & 75.82 \\
\cellcolor{white} & \cellcolor{white} \multirow{-2}{*}{PU} & QAT & 0.48 & 3.89 & 15.68 & 70.28 \\
& \cellcolor{blue!15} & PTQ & 0.03 & 3.38 & 5.05 & 70.43 \\ 
\cellcolor{white} \multirow{-4}{*}{SSRN} & \multirow{-2}{*}{SA} & QAT & 0.33 & 3.45 & 18.27 & 70.72 \\
\hline
\cellcolor{blue!15} & & PTQ & 0.08 & 3.66 & 8.43 & 67.77 \\ 
& \cellcolor{white}\multirow{-2}{*}{PU} & QAT & 0.43 & 3.84 & 10.92 & 67.52 \\ 
\cellcolor{blue!15} & \cellcolor{blue!15} & PTQ & 0.02 & 2.9 & 3.07 & 67.56 \\ 
\multirow{-4}{*}{hybridSN} & \multirow{-2}{*}{SA} & QAT & 0.25 & 2.85 & 9.96 & 68.18 \\ 
\end{tabular}\\
{\raggedright\textcolor{Blue}{\footnotesize All results have a p-value $< 3.3 * 10^{-4}$ and an effect size $> 0.98$ (large)}}
\end{table}

\textit{\textbf{Population-based metaheuristic algorithms vs Random Sampling: }} Table \ref{tab:table_4} demonstrates a comparison of the median of DiR and VR of all seeds between RS and DiverGet. Although random sampling from our synthetic inputs enables the detection of behavioral deviations, the revealed occurrences remain very low compared to search-based approach (using GA or PSO) due to the high-dimensional search space and the lack of optimization over generations. 

\begin{tcolorbox}[colback=blue!5,colframe=blue!40!black]
\textbf{Finding 2:} DiverGet's searching strategy using population-based metaheuristic succeed in outperforming the Random Sampling strategy into steering the generation into prominent regions.
\end{tcolorbox}


\subsection{RQ2: DiverGet Facing Multiple Settings}

\textbf{Motivation:} Throughout its quantization assessment steps, DiverGet supports the instantiation with multiple settings that influence its effectiveness regarding different quality aspects of the generation process of difference-inducing data transformations.

\textbf{Method:} We experiment DiverGet on (i) original test samples from both HSIs, (ii) state-of-the-art DNNs, SSRN and HybridSN, and (iii) conventional 8-bit quantization methods, PTQ and QAT. 
We worked with 50 random seeds from a pool of 8000 original inputs.
All the original test inputs are sampled from the subset of test datasets, for which the original model correctly predicts their corresponding labels. Regarding the settings of DiverGet, we try different options for each experiment. We vary the selected metaheuristic algorithms, either PSO or GA, as well as the targeted fitness functions, $f^{div}$ and $f^{cov}$. Throughout the trials, we keep track of (i)  the divergence rate (DiR) and (ii) the Validation Rate (VR).

\textbf{Results:} Tables \ref{tab:table_5}, and \ref{tab:table_6}, along figures \ref{fig:table_7}, and \ref{fig:table_8} report the average of the selected measurements (i.e., DiR and VR) with respect to one or multiple controlled variables of our experiments (i.e., models, datasets, quantization techniques, fitness functions, and metaheuristics).(more detailed results can be found in the supplementary materials)

\begin{table}[!h]
\centering
\caption{Comparison of the median values of DiR and VR when using DiverGet with $f^{cov}$ and $f^{div}$}
\label{tab:table_5}
\rowcolors{1}{blue!15}{}
\renewcommand{\arraystretch}{1.5}
\arrayrulecolor{Blue}
\begin{tabular}{ccllll}
\rowcolor{Blue}
 &  & \multicolumn{2}{c}{\textcolor{white}{$f^{div}$}} & \multicolumn{2}{c}{\textcolor{white}{$f^{cov}$}}  \\
 \rowcolor{Blue} 
\multirow{-2}{*}{\textcolor{white}{Data}} & \multirow{-2}{*}{\textcolor{white}{Quantization}} & \multicolumn{1}{c}{\textcolor{white}{DiR}} & \multicolumn{1}{c}{\textcolor{white}{VR}}  & \multicolumn{1}{c}{\textcolor{white}{DiR}} & \multicolumn{1}{c}{\textcolor{white}{VR}}   \\
 & PTQ & \multicolumn{1}{r}{22.47(*)} & \multicolumn{1}{r}{72.36(**)}  & \multicolumn{1}{r}{9.57(*)} & \multicolumn{1}{r}{70.74(**)}   \\
\multirow{-2}{*}{\cellcolor{white}PU} & QAT & 21.15(*) & 71.57(*)  & 7.37(*) & 68.02(*)  \\
\hline
\cellcolor{blue!15} & PTQ & 8.18(*) & 70.54(**)  & 0.99(*) & 69.29(**)  \\ 
\multirow{-2}{*}{\cellcolor{blue!15}SA} & QAT & 18.17(*) & 71.02(*)  & 9.80(*) & 68.03(*)  \\
\end{tabular}\\
{\raggedright\textcolor{Blue}{\footnotesize All results have a p-value $< 7.7 * 10^{-4}$.}}\\
{\raggedright\textcolor{Blue}{\footnotesize Effect sizes indicated as: * $>0.918$(large), ** $0.69-0.71$ (medium)}}
\end{table}

\begin{figure}[!h]
  \centering
  \includegraphics[scale=0.75]{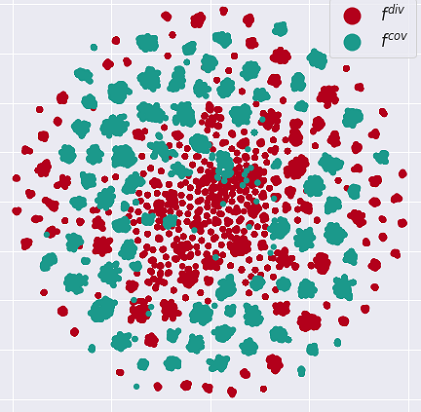}
  \caption{t-SNE visualization of transformation vectors induced by $f^{div}$ and $f^{cov}$.}
  \label{fig:t-SNE}
\end{figure}

\begin{figure}[!h]
  \centering
  \includegraphics[width=1.0\textwidth]{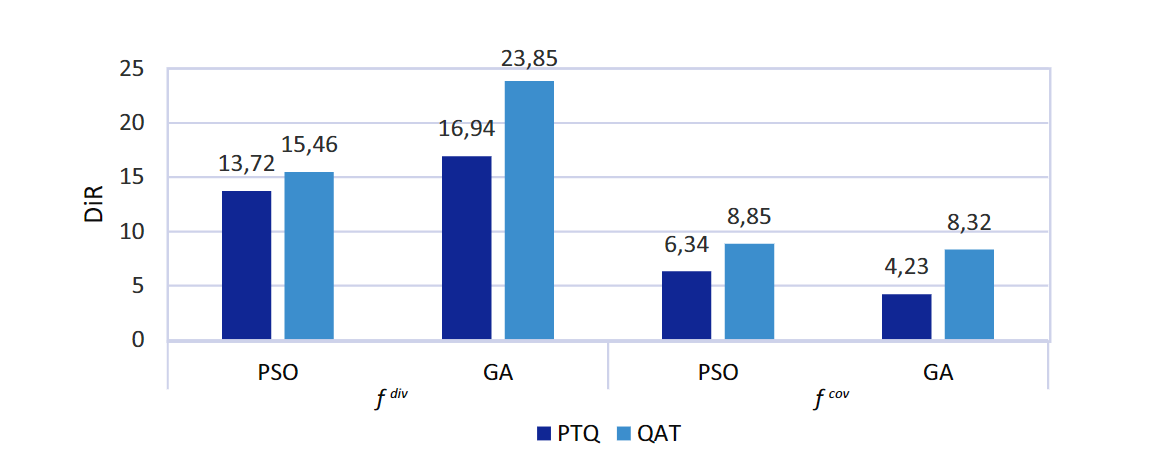}
  \caption{Comparison of the median values of DiR between DiverGet implementations using QAT-based and PTQ-based quantization models. All results have a p-value $< 5.8 * 10^{-9}$ and an effect size $> 0.84$ (large).}
  \label{fig:table_7}
\end{figure}

\begin{figure}[!h]
  \centering
  \includegraphics[width=1.0\textwidth]{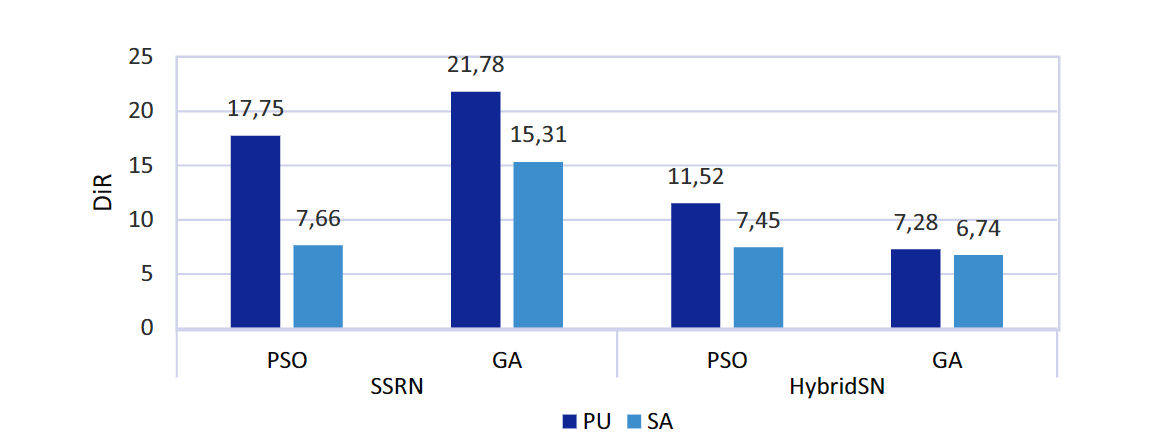}
  \caption{Comparison of the median values of DiR between DiverGet implementations using different datasets. All results have a p-value $< 1.5 * 10^{-11}$ and an effect size $> 0.89$ (large).}
  \label{fig:table_8}
\end{figure}

\textit{\textbf{Divergence-based fitness vs coverage-based fitness: }}Table \ref{tab:table_5} presents the comparison of DIIs and VRs when we use metaheuristic searching with either coverage-based fitness, $f^{cov}$, or divergence-based fitness, $f^{div}$. Results show that $f^{div}$ outperforms $f^{cov}$ in guiding the generation of data transformation towards prominent regions. This is aligned with our expectations given that the sharp guidance of $f^{div}$ directly targets the enlargement of the deviation between the two versions of the DNN. However, we discussed the mode collapse risk when the search process was primarily rewarded for the detection of deviations without any implicit or explicit incentives on their diversification. In the following, we aim to get more insights on the distribution of the transformation vectors that have been discovered by both search strategies. We project these vectors into a two-dimensional sub-space using t-Distributed Stochastic Neighbor Embedding (t-SNE)~\citep{van2008visualizing} to visualize the generated transformation distribution. t-SNE, a nonlinear dimensionality reduction method, embeds the transformation vectors into a lower-dimensional space by generating probability distributions that preserve the similarity of high dimensional data. Figure \ref{fig:t-SNE} presents the outcome of this projection. It shows that $f^{cov}$ tends to discover more separate regions, to which difference-inducing inputs belong. In contrast, $f^{div}$ focuses on a particular central region with a high density of difference-inducing inputs, but it has not prevented the metaheuristic search from arbitrarily trying more distant transformations with fewer similarities that are quite dispersed in the search space.

To support our statement, we performed a statistical analysis on these randomly sampled transformation vectors. More specifically, we measured the pairwise Euclidean distances in each group (i.e, the group generated using $f^{cov}$ and $f^{div}$) of more than $10000$ transformation vectors (the experience was repeated $3 times$). We then reported the average minimum, maximum, mean, variance, and first and third quartile of the two group distances. Table \ref{tab:table_diversity} presents the obtained results. All the statistical indicators (except the maximum) demonstrate that transformation vectors generated using $f^{cov}$ are more diverse and more spread in the search space than the ones generated using $f^{div}$. These results reinforce the statement that $f^{cov}$ could be an alternative to prevent the mode collapse, since it can be used to enhance the diversity of generated transformation vectors (and ultimately DIIs).

\begin{table}[!h]
\centering
\caption{Statistics over the pairwise distances of transformation vectors generated using $f^{div}$ and $f^{cov}$}
\label{tab:table_diversity}
\rowcolors{1}{blue!15}{}
\renewcommand{\arraystretch}{1.5}
\begin{tabular}{ccccccc}
\rowcolor{Blue} \textcolor{white}{Fitness Function} & \textcolor{white}{Min} & \textcolor{white}{Max} & \textcolor{white}{Mean} & \textcolor{white}{Var} & \textcolor{white}{Q1} & \textcolor{white}{Q3} \\
$f^{div}$ & 0 & 1268.8 & 572.2 & 25770.21 & 461.72 & 671.1 \\
$f^{cov}$ & 0 & 1268.73 & 602.37 & 29791.4 & 486.83 & 712.42 \\
\end{tabular}
\end{table}

\begin{tcolorbox}[colback=blue!5,colframe=blue!40!black]
\textbf{Finding 3:} Divergence-based fitness function succeeds in steering the input generation towards regions of higher density of disagreement-revealing ability than its coverage-based counterpart, while preserving the test data diversification.
\end{tcolorbox}


\textit{\textbf{GA vs PSO as metaheuristic searching algorithm: }}Table \ref{tab:table_6} presents the comparison between PSO and GA as a metaheuristic for DiverGet using the same aforementioned measures. Results shows that PSO and GA succeed in improving the disagreement-revealing ability over the generations. Indeed, our in-depth investigation on each metaheuristic's optimization routines and their exposed transformation vectors, leads us to believe that issues in relation with exploitation/exploration are the main reason behind this performance gap. On the first hand, PSO tends to explore more regions since all particles have to move, even slightly, in each generation. On the other hand, GA tends to highly exploit the most-fitted candidates via breeding between them, to converge quickly towards the best-fitted solution. Hence, GA fails to explore farther candidate transformations in other regions. Besides, GA applies random mutations on the new generations of candidates aiming at reaching distant regions, but the non-determinism of this arbitrary mutation increases the occurrences of out of the space candidates, as evidenced by the obtained low valid rates (VR). For such search-based approach where the exploration has higher priority than smooth and rapid convergence, PSO might be more appropriate metaheuristic searching algorithm.

\begin{table}[!h]
\centering
\caption{Comparison of the median values of DiR and VR between PSO and GA as metaheuristics for DiverGet}
\label{tab:table_6}
\rowcolors{1}{blue!15}{}
\renewcommand{\arraystretch}{1.5}
\arrayrulecolor{Blue}
\begin{tabular}{cccccc}
\rowcolor{Blue} 
\rowcolor{Blue}  &  & \multicolumn{4}{c}{\begin{tabular}[c]{@{}c@{}}\textcolor{white}{DiverGet}\end{tabular}}  \\ 
\rowcolor{Blue}  &  & \multicolumn{2}{c}{\textcolor{white}{PSO}} & \multicolumn{2}{c}{\textcolor{white}{GA}}  \\
\rowcolor{Blue}\multirow{-3}{*}{\textcolor{white}{Data}} & \multirow{-3}{*}{\textcolor{white}{Quantization}} & \multicolumn{1}{c}{\textcolor{white}{DiR}} & \multicolumn{1}{c}{\textcolor{white}{VR}}  & \multicolumn{1}{c}{\textcolor{white}{DiR}} & \multicolumn{1}{c}{\textcolor{white}{VR}}  \\
 & PTQ & \multicolumn{1}{r}{16.28(**)} & \multicolumn{1}{r}{83.04(*)}  & \multicolumn{1}{r}{15.76(**)} & \multicolumn{1}{r}{60.05(*)}   \\ 
\multirow{-2}{*}{\cellcolor{white}PU} & QAT & 12.99(*) & 80.42(*)  & 15.53(*) & 59.18(*)  \\
 \hline
\cellcolor{blue!15} & PTQ & 3.78(*) & 81.04(*)  & 5.40(*) & 58.79(*)  \\ 
\multirow{-2}{*}{SA} & QAT & 11.32(*) & 79.98(*)  & 16.64(*) & 59.07(*)  
\end{tabular}\\
{\raggedright\textcolor{Blue}{\footnotesize * indicates a p-value $< 8.1 * 10^{-13}$ and an effect size $> 0.92$ (large)}}\\
{\raggedright\textcolor{Blue}{\footnotesize ** indicates a p-value $= 0.133$ and an effect size $= 0.59$ (small)}}
\end{table}

\begin{tcolorbox}[colback=blue!5,colframe=blue!40!black]
\textbf{Finding 4:} PSO outperforms GA, as a more adequate metaheuristic-based search algorithm in terms of the size/ratio of its uncovered difference-inducing inputs.
\end{tcolorbox}



\textit{\textbf{PTQ vs QAT as robust quantization technique: }}Figure \ref{fig:table_7} reports the rate of generated difference-inducing inputs when assessing both of QAT-based and PTQ-based quantized models built on the subjects CNN models and HSI datasets. Results show that QAT-based spawn higher rate of behavioral divergences than its PTQ-based counterpart. This is not aligned with the test accuracy evaluations because the QAT is supposed to refine the rounding of floating-point parameters with more precise ranges estimated on the training data. This suggests that the quantization methods that rely strongly on the original data distribution at different levels, including reduction of precision and evaluation, will often be subject to widely divergent behaviors when tested on shifty data inputs that are closer to real contexts and quite dissimilar with the major situations.

\begin{tcolorbox}[colback=blue!5,colframe=blue!40!black]
\textbf{Finding 5:} Quantization-aware training can be less robust than post-training quantization, against naturally-occurring data distortions that can accentuate the gap between synthetic and training data distributions.
\end{tcolorbox}



\textit{\textbf{PU vs SA as complex HSI classification problem: }}Figure \ref{fig:table_8} reports the rate of generated difference-inducing inputs when assessing quantized models that solve two different HSI classification problems. Results highlight that disagreements are more likely to occur using PU than SA. This is mainly due to the nature of each scene. In fact, SA was captured over a rural location comprising large homogeneous fields, which resulted in large and rather uniformly-shaped classes with homogeneous neighborhoods. As a consequence, DNNs can easily classify any 3D patch from SA, even when distorted, which makes generating difference-inducing inputs more challenging. However, PU was captured over an urban site where different land covers are represented by small and close regions that are scattered all over the image. As a result, and given PU's heterogeneous neighborhoods, distorting a 3D patch can harshly affect the spatial information, mislead the classifier, and easily generate more difference-inducing inputs. Last, in line with~\citep{xie_diffchaser:_2019}, larger models like SSRN have higher chances to generate more disagreements than hybridSN (see Figure\ref{fig:table_8}). Indeed, SSRN is a residual network composed of more layers than hybridSN.

\begin{tcolorbox}[colback=blue!5,colframe=blue!40!black]
\textbf{Finding 6:} The complexity of the HSI classification problem, especially in terms of homogeneity of the pixel patch, can make the systematic assessment via naturally-occurring data distortions, more challenging to pass for quantized CNNs. 
\end{tcolorbox}


\subsection{RQ3: DiverGet vs DiffChaser}
\textbf{Motivation: }The goal is to compare DiverGet's performance in discovering difference-inducing inputs in comparison with DiffChaser, the state-of-the-art framework for detecting disagreement among DNNs.

\textbf{Method:} We slightly modified DiffChaser to add support for HSI. We used DiffChaser with its basic fitness function since it is (i) untargeted and (ii) results in more DIIs than the other DiffChaser k-uncertainty fitnesses. For a fair comparison, we chose the \emph{$f^{div}$} as our fitness function since it exploits the last layer of the two DNNs to produce divergence which is quite similar to DiffChaser’s basic fitness function design. We experimented on 50 random seeds from a pool of $8000$ 3D patches from the two test data of SA and PU. We also used the two quantization methods PTQ and QAT with the two HSI models SSRN and HybridSN. To provide the most accurate comparison, we run both frameworks with the same population size, $10$, and with the same number of iterations, $25$, that would yield in both evaluation the total of $10\times25\times8000$ synthetic samples. As measurements, we use the two metrics, divergences rate (DiR) and the success rate (SR), i.e., the percentage of original 3D patches where at least one disagreement is successfully revealed among their descendant synthetic inputs.

In our quantization assessment strategy, DiR estimates the expectation of quantized model divergence rates against natural perturbations by analogy with estimation of the expected model's misclassification. This complements other metrics, such as success rate and time to find first disagreements (used in DiffChaser), that demonstrate the unavoidable existence of divergence between two DNN  versions  with  different  arithmetic-precision. To assess the effectiveness of different quantized models, we compare the likelihood that they will fail when faced with naturally-occurring distorted inputs.

To have a better understanding, we summarize the difference between the two frameworks in Table \ref{tab:table_9}.

\begin{table}[!h]
\centering
\caption{DiverGet vs DiffChaser}
\label{tab:table_9}
\rowcolors{1}{blue!15}{}
\renewcommand{\arraystretch}{1.5}
\begin{tabular}{p{0.11\textwidth}p{0.39\textwidth}p{0.39\textwidth}}
\hline
\rowcolor{Blue} \multicolumn{1}{c}{\textcolor{white}{Level}} & \multicolumn{1}{c}{\textcolor{white}{DiverGet}} & \multicolumn{1}{c}{\textcolor{white}{DiffChaser}} \\
Distortion design & Naturally-occurring perturbation & White noise injection \\
Constraints & PSNR & $L_{\infty}$ norm \\
Search space & Transformation vector space & Input space \\
Search objective & Behavioral Divergence Maximization & Targeted/Untargeted attack \\
Goal & DIIs generation: Exposing every possible divergence for realistic quantization assessment against corner case scenarios. & First disagreement attack: if disagreement is found DiffChaser  stops or re-starts from random state. 
\end{tabular}
\end{table}

\textbf{Results:} Table \ref{tab:table_10} summarizes the overall results. It shows that DiverGet configured with PSO successfully generates disagreements for $40.98\%$ of 3D patches on average, with an average DiR of $14.59\%$. DiverGet can still generate disagreements when applied with GA for an average of $27.27\%$ of 3D patches with a DiR of $20.40\%$. For DiffChaser, the success rate is $11.25\%$ on average, which is lower than DiverGet with both metaheuristics options. Moreover, the gap between the two frameworks is also remarkable in regards to the rates of revealed divergences. In fact, DiffChaser generates an average DiR of $2.78\%$ which is quite low compared to DiverGet.

\begin{table}[!h]
\centering
\caption{Comparison of the median values of DiR and SR between DiffChaser, PSO-based DiverGet and GA-based DiverGet}
\label{tab:table_10}
\rowcolors{1}{blue!15}{}
\renewcommand{\arraystretch}{1.5}
\begin{tabular}{P{0.11\textwidth}P{0.07\textwidth}P{0.039\textwidth}P{0.039\textwidth}P{0.039\textwidth}P{0.039\textwidth}P{0.039\textwidth}P{0.039\textwidth}P{0.039\textwidth}P{0.039\textwidth}P{0.039\textwidth}P{0.039\textwidth}}
\rowcolor{Blue} &  & \multicolumn{4}{c}{\textcolor{white}{PU}} & \multicolumn{4}{c}{\textcolor{white}{SA}}& \multicolumn{2}{c}{}  \\
\rowcolor{Blue} &  & \multicolumn{2}{c}{\textcolor{white}{PTQ}} & \multicolumn{2}{c}{\textcolor{white}{QAT}} & \multicolumn{2}{c}{\textcolor{white}{PTQ}} & \multicolumn{2}{c}{\textcolor{white}{QAT}} &
\multicolumn{2}{c}{\multirow{-2}{*}{\textcolor{white}{Average}}}\\ 
\rowcolor{Blue} \multirow{-3}{*}{\textcolor{white}{Framework}} & \multirow{-3}{*}{\textcolor{white}{Model}} & \textcolor{white}{DiR} & \textcolor{white}{SR} & \textcolor{white}{DiR} & \textcolor{white}{SR} & \textcolor{white}{DiR} & \textcolor{white}{SR} & \textcolor{white}{DiR} & \textcolor{white}{SR} & \textcolor{white}{DiR} & \textcolor{white}{SR} \\
 & SSRN & 16.66 & 49.38 & 0.31 & 10.63 & 0.35 & 9.69 & 3.68 & 16.58 & \cellcolor{white} & \cellcolor{white} \\ 
\cellcolor{white}\multirow{-2}{*}{\cellcolor{white}DiffChaser} &\cellcolor{white} Hybrid -SN &\cellcolor{white}  0.001 &\cellcolor{white}  0.31 &\cellcolor{white}  0.002 &\cellcolor{white}  0.31 &\cellcolor{white}  0.001 &\cellcolor{white}  0.63 &\cellcolor{white}  1.22 &\cellcolor{white}  2.50 & \multirow{-2}{*}{\cellcolor{white}2.78} & \multirow{-2}{*}{\cellcolor{white}11.25} \\
\hline
\multirow{1.5}{*}{\cellcolor{blue!15}DiverGet} &\cellcolor{blue!15} SSRN &\cellcolor{blue!15}  24.96 &\cellcolor{blue!15} 71.25 &\cellcolor{blue!15}  16.42 &\cellcolor{blue!15} 61.25 &\cellcolor{blue!15}  3.60 &\cellcolor{blue!15} 43.75 &\cellcolor{blue!15}  13.92 &\cellcolor{blue!15}  63.44 & \cellcolor{blue!15} & \cellcolor{blue!15} \\ 
\multirow{-1.5}{*}{\cellcolor{blue!15}(PSO)} & Hybrid -SN & 16.97 & 24.38 & 20.08 (**) & 28.75 & 9.35 & 14.06 & 11.42 & 20.94 (*) & \multirow{-2}{*}{\cellcolor{blue!15}14.59} & \multirow{-2}{*}{\cellcolor{blue!15}40.98} \\
\hline
\multirow{1.5}{*}{\cellcolor{white}DiverGet}  & SSRN & 35.90 & 58.75 & 28.86 & 37.50 & 14.47 & 20.63 & 31.93 & 43.75 & \cellcolor{white} & \cellcolor{white} \\ 
\multirow{-1.5}{*}{\cellcolor{white}(GA)} &\cellcolor{white} Hybrid -SN &\cellcolor{white} 12.06 &\cellcolor{white} 13.75 &\cellcolor{white} 19.22 (**) &\cellcolor{white} 20.94 &\cellcolor{white} 5.32 &\cellcolor{white} 5.63 &\cellcolor{white} 15.40 &\cellcolor{white} 17.19 (*) & \multirow{-2.2}{*}{\cellcolor{white}20.40} & \multirow{-2.2}{*}{\cellcolor{white}27.27} \\
\hline
\end{tabular}\\
{\raggedright\textcolor{Blue}{\footnotesize * indicates an effect size $= 0.73$ (medium)}}\\
{\raggedright\textcolor{Blue}{\footnotesize ** indicates a p-value $= 0.305$ and an effect size $= 0.55$ (negligible)}}\\
{\raggedright\textcolor{Blue}{\footnotesize All the remaining results have a p-value $< 4.6 * 10^{-5}$ and an effect size $> 0.82$ (large).}}
\end{table}

This gap may be explained by the following differences: \textit{(i) Search space encoding:} we define the transformation vector space as our search space instead of the input space. This reduces significantly the search space, enables the inclusion of diverse distortion type, and allows a better convergence of the optimizer. HSI classification problem sheds the light on the effect of our variable change' trick in reducing the high-dimensionality of HSI into a single vector of transformation metadata; \textit{(ii) Metamorphic transformations:} DiverGet uses several radiometric and geometric transformations to enrich the transformation space whereas DiffChaser simply applies white noise injection. This offers a great advantage to our approach and allows our search process to explore prominent regions and enhances the chances to expose hidden divergences; \textit{(iii) First-occurrence disclosure objective: }DiffChaser was designed first to focus entirely on exploring the input space with the objective of discovering the first disagreement. This might also explain this performance disparity compared to DiverGet that seeks to produce the most difference-inducing inputs from each 3D patch.

The time required to detect the first disagreement is a primary metric used to evaluate DiffChaser. In this experiment, we are aiming to evaluate DiverGet using this metric and compare it to DiffChaser. To that end, we used for this experiment a subset of 3 seeds (from the 50, previously mentioned, seeds) on which we run DiffChaser and DiverGet. We slightly modified DiverGet to support the first attack concept introduced in DiffChaser and computed for each original HSI the time required to identify the first disagreement per patch (FDI). In addition, as indicated in DiffChaser's publication, we measured the time required to identify the first disagreement per patch only for successful cases (i.e., original HSIs that resulted in DIIs) (FDI*). The intuition behind this measure is that it will report the time without penalizing the framework if it does not identify a disagreement for a particular patch. The outcomes of this experiment are shown in Table \ref{tab:table_11}. The experiment demonstrates that the results are strongly dependent on the model under test. When using SSRN, DiffChaser outperforms DiverGet configured with both metaheuristics. GA-based DiverGet generates the first DII in roughly 66 seconds on average, and 39 seconds when only successful cases are considered (FDI*). DiverGet with PSO configuration performed better. It takes 36 seconds on average to generate the first DII (26 seconds when only successful cases are considered). DiffChaser, on the other hand, is more efficient than DiverGet configured with both metaheuristics in detecting the first disagreement. In fact, given all seed inputs, it takes around 7 seconds to produce the first disagreement and 2 seconds to generate the first disagreement for successful cases only. When hybridSN is used as a model under test, the situation is mirrored. With an average FDI of 14 seconds (and an average FDI* of 5 seconds), DiffChaser becomes the slower framework. DiverGet, based on GA, performs better, with an average FDI of 11 seconds (an average FDI* of 5 seconds). Finally, with an average FDI of 7 seconds, PSO-based DiverGet has the best average FDI.

In summary, based on the two previous experiments DiverGet outperforms DiffChaser in terms of DiR and SR, with its PSO version performing a comparable FDI.

\begin{table}[!h]
\centering
\caption{Comparison of the median values of FDI and FDI* between DiffChaser, PSO-based DiverGet and GA-based DiverGet}
\label{tab:table_11}
\rowcolors{1}{blue!15}{}
\renewcommand{\arraystretch}{1.5}
\begin{tabular}{P{0.07\textwidth}P{0.11\textwidth}P{0.039\textwidth}P{0.039\textwidth}P{0.039\textwidth}P{0.039\textwidth}P{0.039\textwidth}P{0.039\textwidth}P{0.039\textwidth}P{0.039\textwidth}P{0.039\textwidth}P{0.039\textwidth}}
\rowcolor{Blue}
 &  & \multicolumn{4}{c}{\textcolor{white}{PU}} & \multicolumn{4}{c}{\textcolor{white}{SA}} & \multicolumn{2}{c}{} \\
\rowcolor{Blue}
 &  & \multicolumn{2}{c}{\textcolor{white}{PTQ}} & \multicolumn{2}{c}{\textcolor{white}{QAT}} & \multicolumn{2}{c}{\textcolor{white}{PTQ}} & \multicolumn{2}{c}{\textcolor{white}{QAT}} &
\multicolumn{2}{c}{\multirow{-2}{*}{\textcolor{white}{Average}}} \\
\rowcolor{Blue}
\multirow{-3}{*}{\textcolor{white}{Model}} & \multirow{-3}{*}{\textcolor{white}{Framework}} & \textcolor{white}{FDI} & \textcolor{white}{FDI*} & \textcolor{white}{FDI} & \textcolor{white}{FDI*} & \textcolor{white}{FDI} & \textcolor{white}{FDI*} & \textcolor{white}{FDI} & \textcolor{white}{FDI*} & \textcolor{white}{FDI} & \textcolor{white}{FDI*} \\ 
 & DiffChaser & 4.11 & 0.99 & 6.93 & 1.71 & 9.06 & 2.86 & 8.83 & 1.37 & 7.23 & 1.73 \\
\cellcolor{white} & DiverGet (PSO) &  23.31 &  17.2 &  26.9 &  20.23 &  54.69 &  38.35 &  39.52 &  29.91 & 36.11 & 26.42 \\ 
\multirow{-3}{*}{SSRN} & DiverGet (GA) &  52.38 & 32.99 &  53.71 & 29.85 &  80.99 & 49.62 &  75.49 &  43.51 & 65.64 & 38.99 \\
\hline
 & DiffChaser & 14.04 & 12.86 & 13.6 & 1.99 & 13.59 & 6.07 & 12.81 & 0.63 & 13.51 & 5.39 \\ 
\cellcolor{blue!15} Hybrid -SN & DiverGet (PSO) & 7.27 & 6.46 & 7.57 & 5.48 & 6.82 & 5.49 & 6.16 & 4.14 & 6.96 & 5.39 \\ 
 & DiverGet (GA) & 10.59 & 5.29 & 13.18 & 5.92 & 10.34 & 5.67 & 9.3 & 4.62 & 10.85 & 5.38 \\ 
\end{tabular}
\end{table}

\begin{tcolorbox}[colback=blue!5,colframe=blue!40!black]
\textbf{Finding 7:} DiverGet outperforms DiffChaser in terms of number of revealed disagreements with a higher success rate.
\end{tcolorbox}


\section{Threats to Validity}
\label{sec:threats}
In this section, we address the potential threats to the validity of our work and emphasize our prevention measures.

\textbf{Selection of subjects: }The selection of our experimental subjects like dataset, models, and quantization methods, can be a threat to validity. As a mitigation strategy, we use two variants of each evaluation subject and all of the considered variants are: (i) state-of-the-art like the tested CNNs, (ii) widely-used by the community like the HSI datasets, (iii) state-of-the-practice like the quantization techniques. Also, we use official implementations of the tested models and we rely on open-source libraries for quite delicate implementations such as quantization techniques and metaheuristic standard algorithms to prevent potential bugs.

\textbf{Design choices:} The design choices that we made throughout the development process can be a threat. To overcome this threat, we implement two competitive metaheuristics for SBST. Furthermore, we perform a prior phase of hyperparameters tuning to balance their configuration in regards to exploration vs exploitation, in order to be suitable for our designed optimization problem. Concerning \textit{$population\_size$} and \textit{$generation\_number$} that control the total number of generated samples, we also tried multiple combinations and we make sure that they guarantee fair comparisons with their opponents, whether it is just a random sampling or the state-of-the-art DiffChaser. Indeed, we extract the naturally-occurring distortions for HSI from remote sensing white papers and we systematically restrict the range of their parameters with regards to the source information loss (PSNR). However, we consulted HSI experts to confirm our final designed distortions, their parameters' ranges, and the threshold of PSNR adopted in the validity test, because these choices would have an impact on the validity of our results.

\textbf{Generalizability to other domains:} Despite most of similar research works study conventional classification problems such as MNIST~\citep{mnist} and CIFAR10~\citep{cifar10}, we opt for more complex problems, HSI classification, as study cases because of their real-world on-edge deployment challenges such as the curse of dimensionality, the prevalence of external context changes, defect-proneness of the acquisition systems. All these challenges are threatening the natural transition to on-edge devices with the conventional assessment strategy, and sheds the light on the importance of advanced quantization assessment to anticipate the risky behavioral deviations by the DNN at the edge and avoid them. Nevertheless, we describe thoroughly the methodology we follow with our domain expert collaborators to instantiate and set up different ingredients of our approach. Also, DiverGet framework is designed in a modular way to support an easy and adaptive plug-and-play configuration.

\textbf{HSI Domain Selection:} HSI DNNs are being employed at the edge in crucial applications related to climate changes, environmental monitoring and Astronomy. Quantification in this domain is challenged by the curse of dimensionality, the ubiquity of external-environment changes, and the defect-proneness of acquisition systems. These factors reinforce the need for advanced assessment. To the best of our knowledge, no work has addressed quantization assessment in the field of HSI. Hence, we tested DiffChaser, a general state-of-the-art computer vision testing framework that claims to be easily extendable to other domains. Its poor performance with HSI DNNs sparked the design of DiverGet, which focuses on naturally-occurring distortions and search design. 

 
\textbf{A fair comparison to RS, and DiffChaser:} We used the number of generations $G$ as metric to fix the total number of model queries allowed per testing session. For metaheuristics, we compute $G = original\_test\_patches \times population\_size \times nb\_iterations$). For RS $G = original\_test\_patches\times \\nb\_samples$. Using the same original test data, we fix equal $population\_size$ and $nb\_iterations$ for DiverGet (GA\&PSO) and DiffChaser (GA). For RS,  $nb\_samples = population\_size \times nb\_iterations$ to enable same number of queries to all testers. Also, we considered the following points for fair comparison with DiffChaser:
\begin{itemize}
  \item Using SR in our evaluation, a metric used in DiffChaser's paper to compare it with our work,
  \item Using single-mode to have same granularity as DiffChaser,
  \item Using $f^{div}$ that exploits the last layer of the two DNNs to produce divergence which is quite similar to DiffChaser’s 'basic' fitness function design,
  \item Disabling the first attack option of DiffChaser and making it generate every possible DIIs without altering its main logic.
\end{itemize}

\textbf{Dealing with randomness in the search algorithms:} For reliable conclusions when comparing search-based approaches, we conducted experiments that take into account the stochasticity inherent in these search algorithms. Comparisons of individual elements might compromise the validity of such experiments and hinder the validity of their outcomes. To mitigate the effects of randomness, all the results included in the empirical evaluations of our approach are the median values estimated over $50$ ($10$ in RQ1) runs that use independently random seeds. We also used statistical hypothesis testing and effect size measurements to assess the statistical significance of our results.

\section{Conclusion}
\label{sec:cl}
In this paper, we introduced DiverGet, a search-based software testing framework dedicated to the assessment of DNN quantization. It (i) relies on domain-specific metamorphic relations to produce semantically preserving data, (ii) leverages various population-based metaheuristic algorithms for a maximum disclosure of difference-inducing inputs, and (iii) operates with two alternative and complimentary fitness functions to guide the search. Our evaluation on HSI classification problems shows that DiverGet can successfully generate meaningful test inputs that induce disagreement among DNNs of different arithmetic precision. It substantially outperforms the state-of-the-art, DiffChaser, in detecting those quantization-induced divergences. We report our systematic approach for the design of novel domain-specific metamorphic relations. This would help the community to easily adapt DiverGet to assess on-edge quantization side effects in other safety-critical domains.

\begin{acknowledgements}
We acknowledge the support from the following organizations and companies: Fonds de Recherche du Québec (FRQ), Natural Sciences and Engineering Research Council of Canada (NSERC), Canadian Institute for Advanced Research (CIFAR), and Huawei Canada. However, the findings and opinions expressed in this paper are those of the authors and do not necessarily represent or reflect those organizations/companies.
\end{acknowledgements}

%
%


\bibliographystyle{spbasic}
\bibliography{sn-bibliography}

\end{document}